\documentclass[11pt]{article}

\usepackage{epsfig}
\usepackage{theorem}
\usepackage{enumerate}
\usepackage{amsmath}
\usepackage{amssymb}
\usepackage{boxedminipage}
\usepackage{subfigure}
\usepackage{url}
\usepackage{color}
\usepackage{epsfig}

\newcommand{\titlestring}{iBOA: The Incremental Bayesian Optimization Algorithm}
\newcommand{\reportnumber}{2008002}
\newcommand{\shortauthors}{Martin Pelikan, Kumara Sastry, and David E. Goldberg}
\newcommand{\datestring}{January 2008}
\definecolor{myblue}{rgb}{0.165,0.34,0.5}
\date{}

\begin{document}

\begin{titlepage}
\setlength{\parindent}{0pt}
%\vspace*{0.5in}

\noindent
\includegraphics[width=5in]{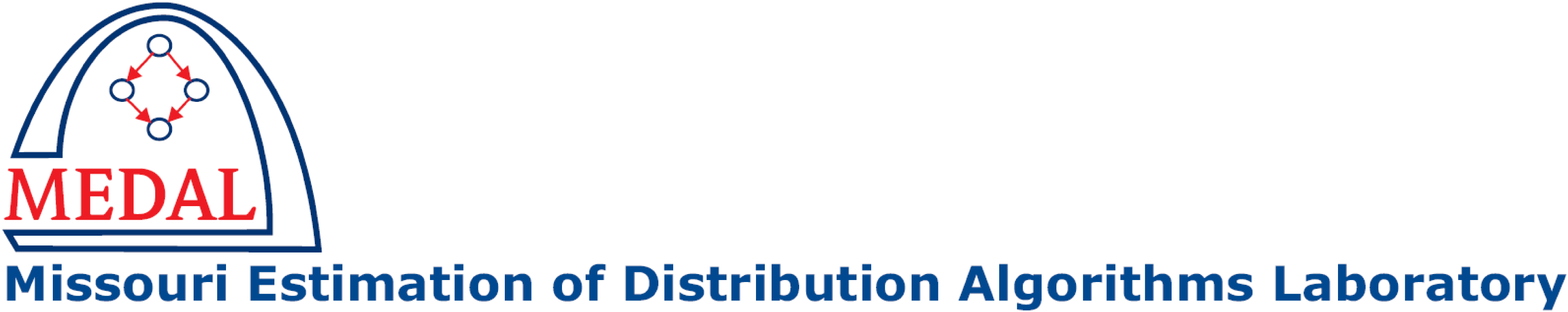}
\vspace*{0.075in}
{\color{myblue}
\hrule height 2pt
}
\vspace*{0.5in}

{\bf
\textsf{{\large
\titlestring}}
}
\vspace*{0.25in}

\textsf{\shortauthors}

\vspace*{0.25in}

\textsf{MEDAL Report No. \reportnumber}

\vspace*{0.25in}

\textsf{\datestring}

\vspace*{0.25in}

{\bf \textsf{Abstract}}  

\vspace*{0.075in}

{\small \textsf{This paper proposes the incremental Bayesian optimization algorithm (iBOA), which modifies standard BOA by removing the population of solutions and using incremental updates of the Bayesian network. iBOA is shown to be able to learn and exploit unrestricted Bayesian networks using incremental techniques for updating both the structure as well as the parameters of the probabilistic model. This represents an important step toward the design of competent incremental estimation of distribution algorithms that can solve difficult nearly decomposable problems scalably and reliably.}}
%\vspace{4in}

\vspace*{0.25in}

{\bf \textsf{Keywords}}

\vspace*{0.075in}
{\small \textsf{Bayesian optimization algorithm, incremental BOA, incremental model update, estimation of distribution algorithms, evolutionary computation.}}

\vfill

\noindent
\begin{minipage}{6in}
%\centering
{\small \textsf{Missouri Estimation of Distribution Algorithms Laboratory (MEDAL)\\
Department of Mathematics and Computer Science\\
University of Missouri--St. Louis\\
One University Blvd.,
St. Louis, MO 63121\\
E-mail: \url{medal@cs.umsl.edu}\\
WWW: \url{http://medal.cs.umsl.edu/}\\}}
\end{minipage}

\end{titlepage}

\title{\titlestring}

\author{
{\bf Martin Pelikan}\\
Missouri Estimation of Distribution Algorithms Laboratory (MEDAL)\\
Dept. of Math and Computer Science, 320 CCB\\
University of Missouri at St. Louis\\
One University Blvd., St. Louis, MO 63121\\
\url{pelikan@cs.umsl.edu}
\and
{\bf Kumara Sastry}\\
Illinois Genetic Algorithms Laboratory (IlliGAL)\\
Department of Industrial and Enterprise Systems Engineering\\
University of Illinois at Urbana-Champaign, Urbana IL 61801\\
\url{kumara@illigal.ge.uiuc.edu}
\and
{\bf David E. Goldberg}\\
Illinois Genetic Algorithms Laboratory (IlliGAL)\\
Department of Industrial and Enterprise Systems Engineering\\
University of Illinois at Urbana-Champaign, Urbana IL 61801\\
\url{deg@uiuc.edu}
}

\maketitle

%==============================================================

\begin{abstract}

\end{abstract}


%==============================================================

\section{Introduction}
Estimation of distribution algorithms (EDAs)~\cite{Baluja:94,Muhlenbein:96**,Larranaga:02,Pelikan:book,Larranaga:06,Pelikan:EDA-book} replace standard variation operators of genetic and evolutionary algorithms by building and sampling probabilistic models of promising candidate solutions. Already some of the earliest estimation of distribution algorithms (EDAs) have completely eliminated the need for maintaining an explicit population of candidate solutions used in most standard evolutionary algorithms, and they updated the probabilistic model incrementally using only a few candidate solutions at a time~\cite{Baluja:94,Harik:98*}. The main advantage of such {\em incremental} EDAs is that memory complexity is greatly reduced. One of the must successful results of this line of research was the application of the compact genetic algorithm (cGA)~\cite{Harik:98*} to a noisy problem with over one billion bits~\cite{Goldberg:07,Sastry:07}. Nonetheless, all incremental EDAs proposed in the past use either univariate models with no interactions between the variables or probabilistic models in the form of a tree. 

This paper proposes an incremental version of the Bayesian optimization algorithm (BOA), which uses Bayesian networks to model promising solutions and sample the new ones. The proposed algorithm is called the {\em incremental} BOA (iBOA). While many of the ideas can be adopted from the work on other incremental EDAs, the design of iBOA poses one unique challenge---how to incrementally update a multivariate probabilistic model without either committing to a highly restricted set of structures at the beginning of the run or having to maintain all possible multivariate statistics that can be useful throughout the run? We propose one solution to this challenge and outline another possible approach to tackling this problem. We then test iBOA on several decomposable problems to verify its robustness and scalability on boundedly difficult decomposable problems. Finally, we outline interesting topics for future work in this area.

The paper starts by discussing related work in section~\ref{section-background}. Section~\ref{section-boa} outlines the standard population-based Bayesian optimization algorithm (BOA). Section~\ref{section-iboa} describes the incremental BOA (iBOA). Section~\ref{section-experiments} presents and discusses experimental results. Section~\ref{section-future-work} outlines some of the most important challenges for future research in this area. Finally, section~\ref{section-conclusions} summarizes and concludes the paper.

%==============================================================

\section{Background}
\label{section-background}
This section reviews some of the incremental estimation of distribution algorithms. Throughout the section, we assume that candidate solutions are represented by fixed-length binary strings, although most of the methods can be defined for fixed-length strings over any finite alphabet in a straightforward manner.

\subsection{Population-Based Incremental Learning (PBIL)}
The population-based incremental learning (PBIL) algorithm~\cite{Baluja:94} was one of the first estimation of distribution algorithms and was mainly inspired by the equilibrium genetic algorithm (EGA)~\cite{Juels:93}. PBIL maintains a probabilistic model of promising solutions in the form of a probability vector. The probability vector considers only univariate probabilities and for each string position it stores the probability of a 1 in that position. For an $n$-bit string, the probability vector is thus a vector of $n$ probabilities $p=(p_1,p_2,\ldots,p_{n})$ where $p_i$ encodes the probability of a 1 in the $i$-th position. Initially, all entries in the probability vector are set to $0.5$, encoding the uniform distribution over all binary strings of length $n$.

In each iteration of PBIL, a fixed number $N$ of binary strings are first generated from the current probability vector; for each new string and each string position $i$, the bit in the $i$th position is set to $1$ with probability $p_i$ from the current probability vector (otherwise the bit is set to 0). The generated solutions are evaluated and $N_{best}$ best solutions are then selected from the new solutions based on the results of the evaluation where $N_{best}<N$. The selected best solutions are then used to update the probability vector. Specifically, for each selected solution $(x_1,x_2,\ldots,x_n)$, the probability vector $p=(p_1,p_2,\ldots,p_n)$ is updated as follows:
\[
p_i \leftarrow p_i (1-\lambda) + x_i \lambda \mbox{~~~~~for all $i\in\{1,\ldots,n\}$},
\]
where $\lambda\in (0,1)$ is the learning rate, typically set to some small value. If $x_i=1$, then $p_i$ is increased; otherwise, $p_i$ is decreased. The rate of increase or decrease depends on the learning rate $\lambda$ and the current value of the corresponding probability-vector entry. In the original work on PBIL~\cite{Baluja:94}, $N=200$, $N_{best}=2$, and $\lambda=0.005$.

Although PBIL does not maintain an explicit population of candidate solutions, the learning rate $\lambda$ can be used in a similar manner as the population-size parameter of standard population-based genetic and evolutionary algorithms. To simulate the effects of larger populations, $\lambda$ should be decreased; to simulate the effects of smaller populations, $\lambda$ should be increased.

\subsection{Compact Genetic Algorithm (cGA)}
The compact genetic algorithm (cGA)~\cite{Harik:98*} also maintains a probability vector instead of a population. Similarly as in PBIL, the initial probability vector corresponds to the uniform distribution over all $n$-bit binary strings and all its entries are thus set to $0.5$. In each iteration, cGA generates two candidate solutions from the current probability vector. Then, the two solutions are evaluated and a tournament is executed between the two solutions. The winner $w=(w_1,\ldots,w_n)$ and the loser $l=(l_1,\ldots,l_n)$ of this tournament are then used to update the probability vector. 

Before presenting the update rule used in cGA, let us discuss the effects of a steady-state update on the univariate probabilities of the probability vector in a population of size $N$ where the winner replaces the loser. If for any position $i$ the winner contains a 1 in this position and the loser contains a 0 in the same position, the probability $p_i$ of a 1 in this position would increase by $1/N$. On the other hand, if the winner contains a 0 in this position and the loser contains a 1, then the probability $p_i$ of a 1 in this position would decrease by $1$. Finally, if the winner and the loser contain the same bit in any position, the probability of a 1 in this position would not change. This update procedure can be simulated even without an explicit population using the following update rule~\cite{Harik:98*}:
\[
p_i 
\begin{cases}
p_i-\frac1N & \mbox{if $w_i=0$ and $l_i=1$}\\
p_i+\frac1N & \mbox{if $w_i=1$ and $l_i=0$}\\
p_i & \mbox{otherwise}
\end{cases}
\]
Although cGA does not maintain an explicit population, the parameter $N$ serves as a replacement of the population-size parameter (similarly as $\lambda$ in PBIL).

Performance of cGA can be expected to be similar to that of PBIL, if both algorithms are set up similarly. Furthermore, cGA should perform similarly to the simple genetic algorithm with uniform crossover with the population size $N$. Even more closely, cGA resembles the univariate marginal distribution algorithm (UMDA)~\cite{Muhlenbein:96**} and the equilibrium genetic algorithm (EGA)~\cite{Juels:93}.

\subsection{EDA with Optimal Dependency Trees}
\label{section-dtEDA}
In the EDA with optimal dependency trees~\cite{Baluja:97a}, dependency-tree models are used and it is thus necessary to maintain not only the univariate probabilities but also the pairwise probabilities for all pairs of string positions. The pairwise probabilities are maintained using an array $A$, which contains a number $A[X_i=a,X_j=b]$ for every pair of variables (string positions) $X_i$ and $X_j$ and every combination of assignments $a$ and $b$ of these variables. $A[X_i=a,X_j=b]$ represents an estimate of the number of solutions with $X_i=a$ and $X_j=b$. Initially, all entries in $A$ are initialized to some constant $C_{init}$; for example, $C_{init}=1000$ may be used~\cite{Baluja:97a}.

Given the array $A$, the marginal probabilities $p(X_i=a,X_j=b)$ are estimated for every pair of variables $X_i$ and $X_j$ and every assignment of these variables as
\[
p(X_i=a,X_j=b) = \frac{A[X_i=a,X_j=b]}{\sum_{a',b'} A[X_i=a',X_j=b']}\cdot
\]
Then, a dependency tree is built that maximizes the mutual information between connected pairs of variables, where the mutual information between any two variables $X_i$ and $X_j$ is given by
\[
I(X_i,X_j) = \sum_{a,b} p(X_i=a,X_j=b) \log \frac{P(X_i=a,X_j=b)}{p(X_i=a)p(X_j=b)}
\]
where marginal probabilities $p(X_i=a)$ and $p(X_j=b)$ are computed from the pairwise probabilities $p(X_i=a,X_j=b)$. The tree may be built using a variant of Prim's algorithm for finding minimum spanning trees~\cite{Prim:57}, minimizing the Kullback-Liebler divergence between the empirical distribution and the dependency-tree model~\cite{Chow:68}. 

New candidate solutions can be generated from the probability distribution encoded by the dependency tree, which is defined as
\[
p(X_1=x_1,\ldots,X_n=x_n) = p(X_r=x_r)\prod_{i\neq r} p(X_i=x_i | X_{p(i)}=x_{p(i)}) 
\]
where $r$ denotes the index of the root of the dependency tree and $p(i)$ denotes the parent of $X_i$. The generation starts at the root, the value of which is generated using its univariate probabilities $P(X_i)$, and then continues down the tree by always generating the variables the parents of which have already been generated. 

Each iteration of the dependency-tree EDA proceeds similarly as in PBIL. First, a dependency tree is built from $A$. Then, $N$ candidate solutions are generated from the current dependency tree and $N_{best}$ best solutions are selected out of the generated candidates based on their evaluation. The selected best solutions are then used to update entries in the array $A$; for each solution $x=(x_1,\ldots,x_n)$, the update rule is executed as follows:
\[
A[X_i=a,X_j=b] \gets 
\begin{cases}
\alpha A[X_i=a,X_j=b] + 1 & \mbox{if $x_i=a$ and $x_j=b$}\\
\alpha A[X_i=a,X_j=b] & \mbox{otherwise}
\end{cases}
\]
where $\alpha\in(0,1)$ is the decay rate; for example, $\alpha = 0.99$~\cite{Baluja:97a}. The above update rule is similar to that used in PBIL. 

While both PBIL and cGA use a probabilistic model based on only univariate probabilities, the EDA with optimal dependency trees is capable of encoding conditional dependencies between some pairs of string positions, enabling the algorithm to efficiently solve some problems that are intractable with PBIL and cGA. Nonetheless, using dependency-tree models is still insufficient to fully cover multivariate dependencies; this may yield the EDA with optimal dependency trees intractable on many decomposable problems with multivariate interactions~\cite{Bosman:99,Pelikan:99a,Goldberg:book,Thierens:99,Pelikan:book}.

%==============================================================

\section{Bayesian Optimization Algorithm (iBOA)}
\label{section-boa}
This section describes the Bayesian optimization algorithm (BOA). First, the basic procedure of BOA is described. Next, the methods for learning and sampling Bayesian networks in BOA are briefly reviewed.

\subsection{Basic BOA Algorithm}
The Bayesian optimization algorithm (BOA)~\cite{Pelikan:98b,Pelikan:99a} evolves a population of candidate solutions represented by fixed-length vectors over a finite alphabet. In this paper we assume that candidate solutions are represented by $n$-bit binary strings, but none of the presented techniques is limited to only the binary alphabet. The first population of candidate solutions is typically generated at random according to the uniform distribution over all possible strings.

\begin{figure}
\centering
\epsfig{file=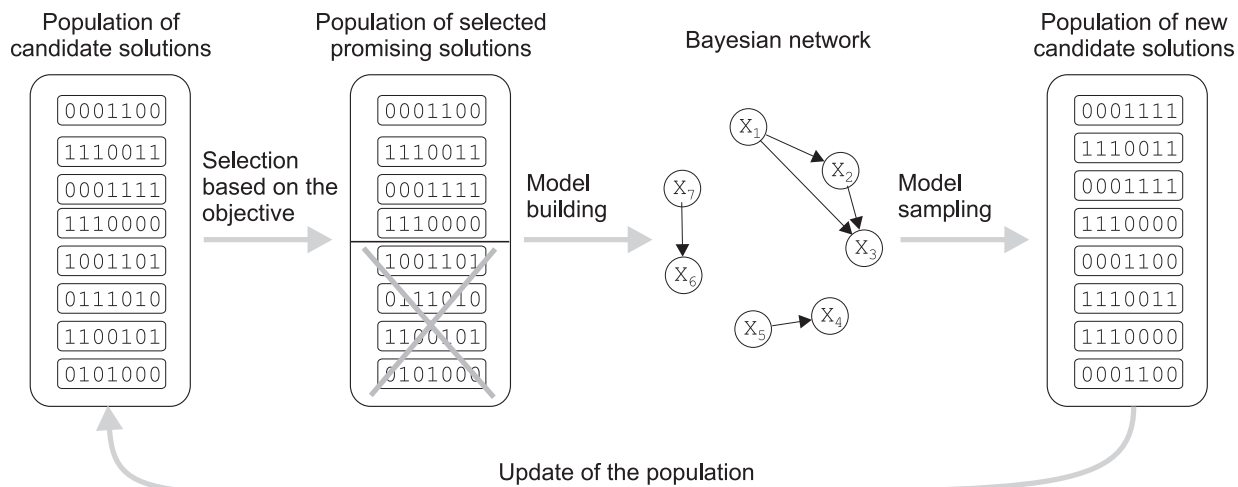,width=0.99\textwidth}
\caption{Basic procedure of the Bayesian optimization algorithm (BOA).}
\label{fig-boa-algorithm}
\end{figure}

Each iteration of BOA starts by selecting a population of promising candidate solutions from the current population. Any selection method used in population-based evolutionary algorithms can be used; for example, we can use binary tournament selection. Then, a Bayesian network is built for the selected solutions. New solutions are generated by sampling the probability distribution encoded by the learned Bayesian network. Finally, the new solutions are incorporated into the original population; for example, this can be done replacing the entire old population with the new solutions. The procedure is terminated when some predefined termination criteria are reached; for example, when a solution of sufficient quality has been reached or when the population has lost diversity and it is unlikely that BOA will reach a better solution than the solution that has been found already. The procedure of BOA is visualized in figure~\ref{fig-boa-algorithm}.

\subsection{Bayesian Networks}
A Bayesian network (BNs)~\cite{Pearl:88,Howard:81} is defined by two components:
\begin{description}
\item[Structure.]~The structure of a Bayesian network for $n$ random variables is defined by an undirected acyclic graph where each node corresponds to one random variable and each edge defines a direct conditional dependency between the connected variables. The subset of nodes from which there exists an edge to the node are called the parents of this node.
\item[Parameters.]~Parameters of a Bayesian network define conditional probabilities of all values of each variable given any combination of values of the parents of this variable.
\end{description}

\noindent
A Bayesian network defines the joint probability distribution
\[
p(X_1,\ldots,X_n) = \prod_{i=1}^n p(X_i|\Pi_i),
\]
where $\Pi_i$ are the parents of $X_1$ and $p(X_i|\Pi_i)$ is the conditional probability of $X_i$ given $\Pi_i$. Each variable directly depends on its parents. On the other hand, the network encodes many independence assumptions that may simplify the joint probability distribution significantly. 

Bayesian networks are more complex than decision trees discussed in section~\ref{section-dtEDA}, allowing BOA to encode arbitrary multivariate dependencies. The estimation of Bayesian networks algorithm (EBNA)~\cite{Etxeberria:99} and the learning factorized distribution algorithm (LFDA)~\cite{Muhlenbein:99} are also EDAs based on Bayesian network models.

\subsection{Learning Bayesian Networks in BOA}
To learn a Bayesian network from the set of selected solutions, we must learn both the structure of the network as well as the parameters (conditional and marginal probabilities). 

To learn the parameters, the maximum likelihood estimation defines the conditional probability that $X_i=x_i$ given that the parents are set as $\Pi_i=\pi_i$ where $x_i$ and $\pi_i$ denote any assignment of the variable and its parents:
\[
p(X_i=x_i|\Pi_i=\pi_i) = \frac{m(X_i=x_i,\Pi_i=\pi_i)}{m(\Pi_i=\pi_i)},
\]
where $m(X_i=x_i,\Pi_i=\pi_i)$ denotes the number of instances with $X_i=x_i$ and $\Pi_i=\pi_i$, and $m(\Pi_i=\pi_i)$ denotes the number of instances with $\Pi_i=\pi_i$.

To learn the structure of a Bayesian network, a greedy algorithm~\cite{Heckerman+al:94} is typically used. In the greedy algorithm for network construction, the network is initialized to an empty network with no edges. Then, in each iteration, an edge that improves the quality of the network the most is added until the network cannot be further improved or other user-specified termination criteria are satisfied.

There are several approaches to evaluating the quality of a specific network structure. In this work, we use the Bayesian information criterion (BIC)~\cite{Schwarz78a} to score network structures. BIC is a two-part minimum description length metric~\cite{Grunwald:98}, where one part represents model accuracy, whereas the other part represents model complexity measured by the number of bits required to store model parameters. For simplicity, let us assume that the solutions are binary strings of fixed length $n$. BIC assigns the network structure $B$ a score~\cite{Schwarz78a} 
\[
\label{equation-BIC}
BIC(B) = \sum_{i=1}^n \left( - H(X_i|\Pi_i) N - 2^{|\Pi_i|} \frac{\log_2(N)}{2} \right),
\]
where $H(X_i|\Pi_i)$ is the conditional entropy of $X_i$ given its parents $\Pi_i$, $n$ is the number of variables, and $N$ is the population size (the size of the training data set). The conditional entropy $H(X_i|\Pi_i)$ is given by
\[
H(X_i|\Pi_i) = -\sum_{x_i,\pi_i} p(X_i=x_i,\Pi_i=\pi_i) \log_2 p(X_i=x_i|\Pi_i=\pi_i),
\]
where the sum runs over all instances of $X_i$ and $\Pi_i$. 

\subsection{Sampling Bayesian Networks in BOA}
\label{section-sampling-boa}
The sampling can be done using the probabilistic logic sampling\index{sampling Bayesian networks!forward simulation} of Bayesian networks~\cite{Henrion:88}, which proceeds in two steps. The first step computes an ancestral ordering of the nodes, where each node is preceded by its parents. 

In the second step, the values of all variables of a new candidate solution are generated according to the computed ordering. Since the algorithm generates the variables according to the ancestral ordering, when the algorithm attempts to generate the value of a variable, the parents of this variable must have already been generated. Given the values of the parents of a variable, the distribution of the values of the variable is given by the corresponding conditional probabilities. 

%==============================================================

\section{Incremental BOA (iBOA)}
\label{section-iboa}
This section outlines iBOA. First, the basic procedure of iBOA is briefly outline. Next, the procedures used to update the structure and parameters of the model are described and it is discussed how to combine these components. Finally, the benefits and costs of using iBOA are analyzed briefly.

\subsection{iBOA: Basic Procedure}
The basic procedure of iBOA is similar to that of other incremental EDAs. The model is initialized to the probability vector that encodes the uniform distribution over all binary strings; all entries in the probability vector are thus initialized to $0.5$. 

\begin{figure}
\centering
\epsfig{file=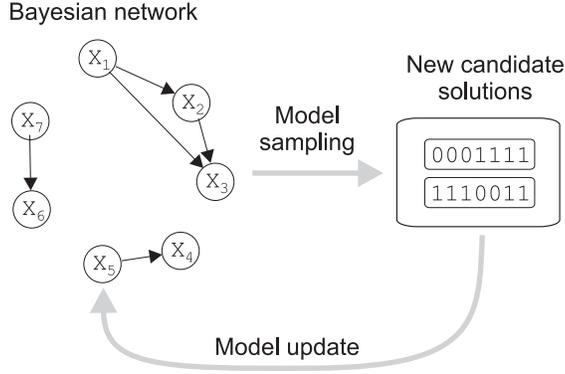,width=0.45\textwidth}
\caption{Basic procedure of the incremental Bayesian optimization algorithm (iBOA).}
\label{fig-iboa-algorithm}
\end{figure}

In each iteration, several solutions are generated from the current model. Then, the generated solutions are evaluated. Given the results of the evaluation, the best and the worst solution out of the generated set of solutions are selected (the winner and the loser). The winner and the loser are used to update the parameters of the model. In some iterations, model structure is updated as well to reflect new dependencies that are supported by the results of the previous tournaments. The basic iBOA procedure is visualized in figure~\ref{fig-iboa-algorithm}.

There are two main differences between BOA and iBOA in the way the model is updated. First of all, the parameters must be updated incrementally because iBOA does not maintain an explicit population of solutions. Second, the model structure also has to be updated incrementally without using a population of strings to learn the structure from. The remainder of this section discusses the details of iBOA procedure. Specifically, we discuss the challenges that must be addressed in the design of the incremental version of BOA. Then, we present several approaches to dealing with these challenges and detail the most important iBOA components.  

\subsection{Updating Parameters in iBOA}
The parameter updates are done similarly as in cGA. Specifically, iBOA maintains an array of marginal probabilities for each string position given other positions that the variable depends on or that the variable {\em may} depend on. Let us denote the winner of the tournament (best solution) by $w=(w_1,\ldots,w_n)$ and the loser (worst solution) by $l=(l_1,\ldots,l_n)$. A marginal probability $p(X_{\beta(1)}=x_{\beta(1)},\ldots,X_{\beta_{k}}=x_{\beta(k)})$ of order $k$ with the positions specified by $\beta(\cdot)$, denoted by by $p(x_{\beta(1)},\ldots,x_{\beta(k)})$ for brevity, is updated as follows:
\[
p(x_{\beta(1)},\ldots,x_{\beta(k)})
\gets
\begin{cases}
p(x_{\beta(1)},\ldots,x_{\beta(k)})
 +\frac1N & \mbox{if $\forall j: w_{\beta(j)}=x_{\beta(j)}$ and $\exists j: l_{\beta(j)}\neq x_{\beta(j)}$}\\
p(x_{\beta(1)},\ldots,x_{\beta(k)})
 -\frac1N & \mbox{if $\exists j: w_{\beta(j)}\neq x_{\beta(j)}$ and $\forall j: l_{\beta(j)}=x_{\beta(j)}$}\\
p(x_{\beta(1)},\ldots,x_{\beta(k)})
& \mbox{otherwise}
\end{cases}
\]

The above update rule increases each marginal probability by $1/N$ if the specific instance is consistent with the winner of the tournament but it is inconsistent with the loser. On the other hand, if the instance is consistent with the loser but not with the winner, the probability is decreased by $1/N$. This corresponds to replacing the winner by the loser in a population of $N$ candidate solutions. For any subset of variables, at most two marginal probabilities change in each update because we only change marginal probabilities for the assignments consistent with either the winner or the loser of the tournament. See figure~\ref{fig-iboa-parameter-update} for an example of the iBOA update rule for marginal probabilities. 

\begin{figure}
\begin{center}
\begin{minipage}{3.5in}
\begin{tabular}{|ccc|c|c|}\hline
&& & $p(X_1,X_3,X_5)$ & $p(X_1,X_3,X_5)$\\
$X_1$ & $X_3$ & $X_5$& (before) & (after)\\\hline
0 & 0 & 0 & 0.10 & 0.10\\
0 & 0 & 1 & 0.05 & {\bf 0.04}\\
0 & 1 & 0 & 0.20 & 0.20\\
0 & 1 & 1 & 0.10 & 0.10\\
1 & 0 & 0 & 0.15 & 0.15\\
1 & 0 & 1 & 0.20 & 0.20\\
1 & 1 & 0 & 0.05 & 0.05\\
1 & 1 & 1 & 0.15 & {\bf 0.16}\\\hline
\end{tabular}
\end{minipage}
\begin{minipage}{2in}
{\bf Population size:}\\
\hspace*{1.5em}$N=100$\\~\\
{\bf Tournament results:}\\
\hspace*{1.5em}{\tt winner~= 101110}\\
\hspace*{1.5em}{\tt loser~~= 010011}\\
\vspace*{3em}
\end{minipage}
\end{center}

\caption{Updating parameters in iBOA proceeds by adding $1/N$ to the marginal probability consistent with the winner and subtracting $1/N$ from the marginal probability consistent with the loser. When the winner and the loser both point to the same entry in the probability table, the table remains the same. }
\label{fig-iboa-parameter-update}
\end{figure}

The conditional probabilities can be computed from the marginal ones. Thus, with the update rule for the marginal probabilities, iBOA can maintain any marginal and conditional probabilities necessary for sampling and structural updates. 

While it is straightforward to initialize any marginal probability under the assumption of the uniform distribution and to update the marginal probabilities using the results of a tournament, one question remains open---what marginal probabilities do we actually need to maintain when we do not know how the model structures will look a priori? Since this question is closely related to structural updates in iBOA, we discuss it next.

\subsection{Updating Model Structure in iBOA}
In all incremental EDAs proposed in the past, already at the beginning of the run it is clear what probabilities have to be maintained. In cGA and PBIL, the only probabilities we have to maintain are the univariate probabilities for different string positions. In the dependency-tree EDA, we also have to maintain pairwise probabilities. But what probabilities do we need to maintain in iBOA? This issue poses a difficult challenge because we do not know model structure a priori and that is why it is not clear what conditional and marginal probabilities we will need. 

Let us first focus on structural updates and assume that the current model is the probability vector (Bayesian network with no edges). To decide on adding the first edge $X_i\rightarrow X_j$ based on the BIC metric or any other standard scoring metric, we need to have pairwise marginal probabilities $p(X_i,X_j)$. In general, let us consider a string position $X_i$ with the set of parents $\Pi_i$. To decide on adding another parent $X_j$ to the current set of parents of $X_i$ using the BIC metric or any other standard scoring metric, we also need probabilities $p(X_i,X_j,\Pi_i)$. 
%If we know the parents for any variable, it is possible to maintain all these marginal probability tables using the update rule presented earlier.

If we knew the current set of parents of each variable, to evaluate all possible edge additions, for each variable, we would need at most $(n-1)$ marginal probability tables. Overall, this would result in at most $n(n-1)=O(n^2)$ marginal probability tables to maintain. However, since we do not know what the set of parents of any variable will be, even if we restricted iBOA to contain at most $k$ parents for any variable, to consider all possible models and all possible marginal probabilities, we would need to maintain at least $n\choose k$ marginal probability tables. Of course, maintaining $n\choose k$ probability tables for relatively large $n$ will be intractable even for moderate values of $k$. This raises an important question---can we do better and store a more limited set of probability tables without sacrificing model-building capabilities of iBOA? 

To tackle this challenge, for each variable $X_i$, we are going to maintain several probability tables. First of all, for any variable $X_i$, we will maintain the probability table for $p(X_i,\Pi_i)$, which is necessary for specifying the conditional probabilities in the current model. Additionally, for $X_i$, we will maintain probability tables $p(X_i,X_j,\Pi_i)$ for all $X_j$ that can become parents of $X_i$, which are necessary for adding a new parent to $X_i$. This will provide iBOA not only with the probabilities required to sample new solutions, but also those required to make a new edge addition ending in an arbitrary node of the network. Overall, the number of subsets for which the probability table will be maintained will be upper bounded by $O(n^2)$, which is a significant reduction from $\Omega(n^{k+1})$ for any $k\geq 2$. 

Nonetheless, we still must resolve the problem of adding new marginal probabilities once we make an edge addition. Specifically, if we add an edge $X_j\rightarrow X_i$, to add another edge ending in $X_i$, we will need to store probabilities $p(X_i,X_j,X_k,\Pi_i)$ where $k$ denotes the index of any other variable that can be added as a parent of $X_i$. While it is impossible to obtain an exact value of these probabilities unless we would maintain them from the beginning of the run, one way to estimate these parameters is to assume independence of $X_k$ and $(X_i,X_j,\Pi_i)$, resulting in the following rule to initialize the new marginal probabilities:
\begin{equation}
\label{eq-initialize-marginal}
p(X_i,X_j,X_k,\Pi_i) = p(X_k) p(X_i,X_j,\Pi_i).
\end{equation}

Once the new marginal probabilities are initialized, they can be updated after each new tournament using the update rule presented earlier. Although the above independence assumption may not hold in general, if the edge $X_k\rightarrow X_i$ is supported by future instances of iBOA, the edge will be eventually added. While other approaches to dealing with the challenge of introducing new marginal probabilities are possible, we believe that the strategy presented above should provide robust performance as is also supported by the experiments presented later. At the same time, after adding an edge $X_j\rightarrow X_i$, we can eliminate probabilities $p(X_i,\Pi_i)$ from the set of probability tables, because this probability table will not be necessary anymore.

Initially, when the model contains no edges, the marginal probabilities $p(X_i,X_j)$ for all pairs of variables $X_i$ and $X_j$ must be stored for the first round of structural update and updated after each tournament. Later in the run, the marginal probabilities for each variable will be changed based on the structure of the model and the results of the tournaments. 

An example sequence of model updates with the corresponding sets of marginal probability tables for a simple problem of $n=5$ bits is shown in figure~\ref{fig-example-probabilities}. Each time a new marginal probability table is added, its values are initialized according to equation~\ref{eq-initialize-marginal}.

\begin{figure}
Initial model: $X_1, X_2, X_3, X_4, X_5$\\
Probabilities stored:\\
\begin{tabular}{ll}
$p(X_1), p(X_2), p(X_3), p(X_4), p(X_5),$ & (for current structure)\\ 
$p(X_1,X_2), p(X_1,X_3), p(X_1,X_4), p(X_1,X_5),$ & (for new parents of $X_1$)\\ 
$p(X_2,X_1), p(X_2,X_3), p(X_2,X_4), p(X_2,X_5),$& (for new parents of $X_2$)\\
$p(X_3,X_1), p(X_3,X_2), p(X_3,X_4), p(X_3,X_5),$& (for new parents of $X_3$)\\
$p(X_4,X_1), p(X_4,X_2), p(X_4,X_3), p(X_4,X_5)$& (for new parents of $X_4$)\\
$p(X_5,X_1), p(X_5,X_2), p(X_5,X_3), p(X_5,X_4)$& (for new parents of $X_5$)\\
\end{tabular}\\
~\\
Added edge: $X_3\rightarrow X_2$\\
New model: $X_1, X_2\leftarrow X_3, X_3, X_4, X_5$\\
Probabilities stored:\\
\begin{tabular}{ll}
$p(X_1), p(X_2, X_3), p(X_3), p(X_4), p(X_5),$ & (for current structure)\\
$p(X_1,X_2), p(X_1,X_3), p(X_1,X_4), p(X_1,X_5),$& (for new parents of $X_1$)\\ 
$p(X_2,X_3, X_1), p(X_2,X_3, X_4), p(X_2,X_3,X_5),$& (for new parents of $X_2$)\\
$p(X_3,X_1), p(X_3,X_4), p(X_3,X_5),$& (for new parents of $X_3$)\\
$p(X_4,X_1), p(X_4,X_2), p(X_4,X_3), p(X_4,X_5)$& (for new parents of $X_4$)\\
$p(X_5,X_1), p(X_5,X_2), p(X_5,X_3), p(X_5,X_4)$& (for new parents of $X_5$)\\
\end{tabular}\\
~\\
Added edge: $X_2\rightarrow X_1$\\
New model: $X_1 \leftarrow X_2, X_2\leftarrow X_3, X_3, X_4, X_5$\\
Probabilities stored:\\
\begin{tabular}{ll}
$p(X_1, X_2), p(X_2, X_3), p(X_3), p(X_4), p(X_5),$ & (for current structure)\\
$p(X_1,X_2, X_3), p(X_1,X_2,X_4), p(X_1,X_2,X_5),$& (for new parents of $X_1$)\\ 
$p(X_2,X_3, X_4), p(X_2,X_3,X_5),$& (for new parents of $X_2$)\\
$p(X_3,X_4), p(X_3,X_5),$& (for new parents of $X_3$)\\
$p(X_4,X_1), p(X_4,X_2), p(X_4,X_3), p(X_4,X_5)$& (for new parents of $X_4$)\\
$p(X_5,X_1), p(X_5,X_2), p(X_5,X_3), p(X_5,X_4)$& (for new parents of $X_5$)\\
\end{tabular}
\caption{iBOA stores all marginal probabilities for the current structure and those required for adding a new edge into any node. Since some edges are disallowed (due to cycles), some probabilities may be omitted. In the above example, for clarity, some marginal probabilities are repeated (this would not be done in the actual implementation). }
\label{fig-example-probabilities}
\end{figure}

\subsection{Sampling New Solutions}
There is no difference between the Bayesian network learned in BOA and iBOA. Therefore, the same sampling algorithm as in BOA can be used in iBOA. Specifically, the variables are first topologically ordered and for each string, the variables are generated according to the generated ancestral ordering using the conditional probabilities stored in the model as described in section~\ref{section-sampling-boa}.

\subsection{Strategies for Combining Components of iBOA}
There are several strategies for combining all the iBOA components described above together. This section briefly reviews and discusses several of these strategies.

The first approach is to perform continuous updates of both the structure as well as the parameters. After performing each tournament, all probabilities will be updated first, and then the structure will be updated by adding any new edges that lead to an improvement of model quality.

The second approach attempts to simulate BOA somewhat closer by updating the structure only once in $N$ iterations; only the probabilities for sampling new solutions will be updated in every iteration of iBOA. This will significantly reduce the complexity of structural updates and improve the overall efficiency. While the model structure will not be updated as frequently as in the first approach, the structural updates might be more accurate because of forcing the metric to use more data to make an adequate structural update. 

The third approach removes the steady-state component of iBOA and updates both the probabilities for sampling new solutions as well as the model structure only once in every $N$ iterations. That means that until the next structural update, the probability distribution encoded by the current model remains constant, and it is only changed once new edges have been added to the new values after the last $N$ parameter updates.

All above approaches can be implemented efficiently, although in practice it appears that the second approach performs the best. The basic procedure of iBOA is outlined in figure~\ref{fig-iboa-pseudocode}.

\begin{figure}[t]
\setlength{\fboxsep}{2.5ex}
\begin{verbatim}
Incremental BOA (iBOA)
  t := 0;
  B := probability vector (no edges);
  p := marginal probabilities for B assuming uniform distribution;
  while (not done) {
    generate k solutions from B with probabilities p;
    evaluate the generated solutions;
    update p using the new solutions;
    update B using the new p;
    t := t+1;
  };
\end{verbatim}
\caption{Pseudocode of the incremental Bayesian optimization algorithm (iBOA). Model structure is denoted by {\tt B}, marginal probabilities are denoted by {\tt p}. Depending on the variant of iBOA, some structural updates may be skipped.}
\label{fig-iboa-pseudocode}
\end{figure}

\subsection{Benefits and Costs}
Clearly, the main benefit of using iBOA instead of the standard BOA is that iBOA eliminates the population and it will thus reduce the memory requirements of BOA. This can be especially important when solving extremely big and difficult problems where the populations may become very large. iBOA also provides the first incremental EDA capable of maintaining multivariate probabilistic models built with the use of multivariate statistics. 

Nonetheless, eliminating the population size also brings disadvantages. First of all, it becomes difficult to effectively maintain diversity using niching, such as restricted tournament selection, because niching techniques typically require an explicit population of candidate solutions. While it might be possible to design specialized niching techniques that directly promote diversity by modifying the probabilistic model in some way, doing this seems to be far from straightforward. Second, while iBOA reduces memory complexity of BOA by eliminating the population, it still is necessary to store the probabilistic model including all marginal probabilities required to make new edge additions. Since the marginal probability tables may require even more memory than the population itself, the memory savings will not be as significant as in cGA or PBIL. Nonetheless, as discussed in the section on future work, this problem may be alleviated by using local structures in Bayesian networks, such as default tables or decision trees. 

%==============================================================

\section{Experiments}
\label{section-experiments}
This section presents experimental results obtained with iBOA on concatenated traps of order 4 and 5, and compares performance of iBOA to that of the standard BOA. 

\subsection{Test Problems}
To test iBOA, we used two separable problems with fully deceptive subproblems based on the well-known trap function:
\begin{itemize}
\item {\em Trap-4.} 
In trap-4~\cite{Ackley:87b,Deb:91c}, the input string is first partitioned into independent groups of $4$ bits each. This partitioning is unknown to the algorithm and it does not change during the run. A 4-bit fully deceptive trap function is applied to each group of 4 bits and the contributions of all trap functions are added together to form the fitness. The contribution of each group of $4$ bits is computed as
\begin{equation}
trap_4(u) = 
\left\{
\begin{array}{ll}
4 & \mbox{~~if $u=4$} \\
3-u & \mbox{~~otherwise}
\end{array}
\right.,
\end{equation}
where $u$ is the number of $1$s in the input string of $4$ bits. The task is to maximize the function. An $n$-bit trap-4 function has one global optimum in the string of all 1s and $(2^{n/4}-1)$ other local optima. Traps of order 4 necessitate that all bits in each group are treated together, because statistics of lower order are misleading. Since hBOA performance is invariant with respect to the ordering of string positions~\cite{Pelikan:book}, it does not matter how the partitioning into 4-bit groups is done, and thus, to make some of the results easier to understand, we assume that trap partitions are located in contiguous blocks of bits.

\item  {\em Trap-5.} 
In trap-5~\cite{Ackley:87b,Deb:91c}, the input string is also partitioned into independent groups but in this case each partition contains $5$ bits and the contribution of each partition is computed using the trap of order 5:
\begin{equation}
trap_5(u) = 
\left\{
\begin{array}{ll}
5 & \mbox{~~if $u=5$} \\
4-u & \mbox{~~otherwise}
\end{array}
\right.,
\end{equation}
where $u$ is the number of $1$s in the input string of $5$ bits. The task is to maximize the function. An $n$-bit trap-5 function has one global optimum in the string of all 1s and $(2^{n/5}-1)$ other local optima. Traps of order 5 necessitate that all bits in each group are treated together, because statistics of lower order are misleading. 
\end{itemize}

\subsection{Description of Experiments}
Although iBOA does not maintain an explicit population of candidate solutions, it still uses the parameter $N$ which loosely corresponds to the actual population size in the standard, population-based BOA. Thus, while iBOA is population-less, we still need to set an adequate population size to ensure that iBOA finds the global optimum reliably. We used the bisection method~\cite{Sastry:01c,Pelikan:book} to estimate the minimum population size to reliably find the global optimum in 10 out of 10 independent runs. To get more stable results, 10 independent bisection runs were repeated for each problem size and thus the results for each problem size were averaged over 100 successful runs. The number of generations was upper bounded by the number of bits $n$, based on the convergence theory for BOA~\cite{Muhlenbein:93c,Thierens:98,Goldberg:book,Pelikan:book} and preliminary experiments. For iBOA, the number of generations is defined as the ratio of the number of iterations divided by the population size. 

In iBOA, the number of solutions in each tournament was set to $k=4$ based on preliminary experiments, which showed that this value of $k$ performed well. To use selection of similar strength in BOA, we used the tournament selection with tournament size $4$. Although these two methods are not equivalent, they should perform similarly. In both BOA and iBOA, BIC metric was used to evaluate competing network structures in model building and the maximum number of parents was not restricted in any way. In iBOA, the model structure is updated once in every $N$ iterations, while the sampling probabilities are updated in each iteration. Finally, in BOA, the new population of candidate solutions replaces the entire original population; while this setting is not optimal (typically elitist replacement or restricted tournament replacement would perform better~\cite{Pelikan:book,Pelikan:06b}), it was still the method of choice to make the comparison fair because iBOA does not use any elitism or niching either.

Although one of the primary goals of setting up BOA and iBOA was to make these algorithms perform similarly, the comparison of these two algorithms is just a side product of our experiments. The most important goal was to provide empirical support for the ability of iBOA to discover and maintain a multivariate probabilistic model without using an explicit population of candidate solutions. We also tried the original cGA; however, due to the use of the simple model in the form of the probability vector, cGA was not able to solve problems of size $n\geq 20$ even with extremely large populations and these results were thus omitted. 

\subsection{Results}
\begin{figure}[t]
\subfigure[Trap-4]{\epsfig{width=0.49 \textwidth,file=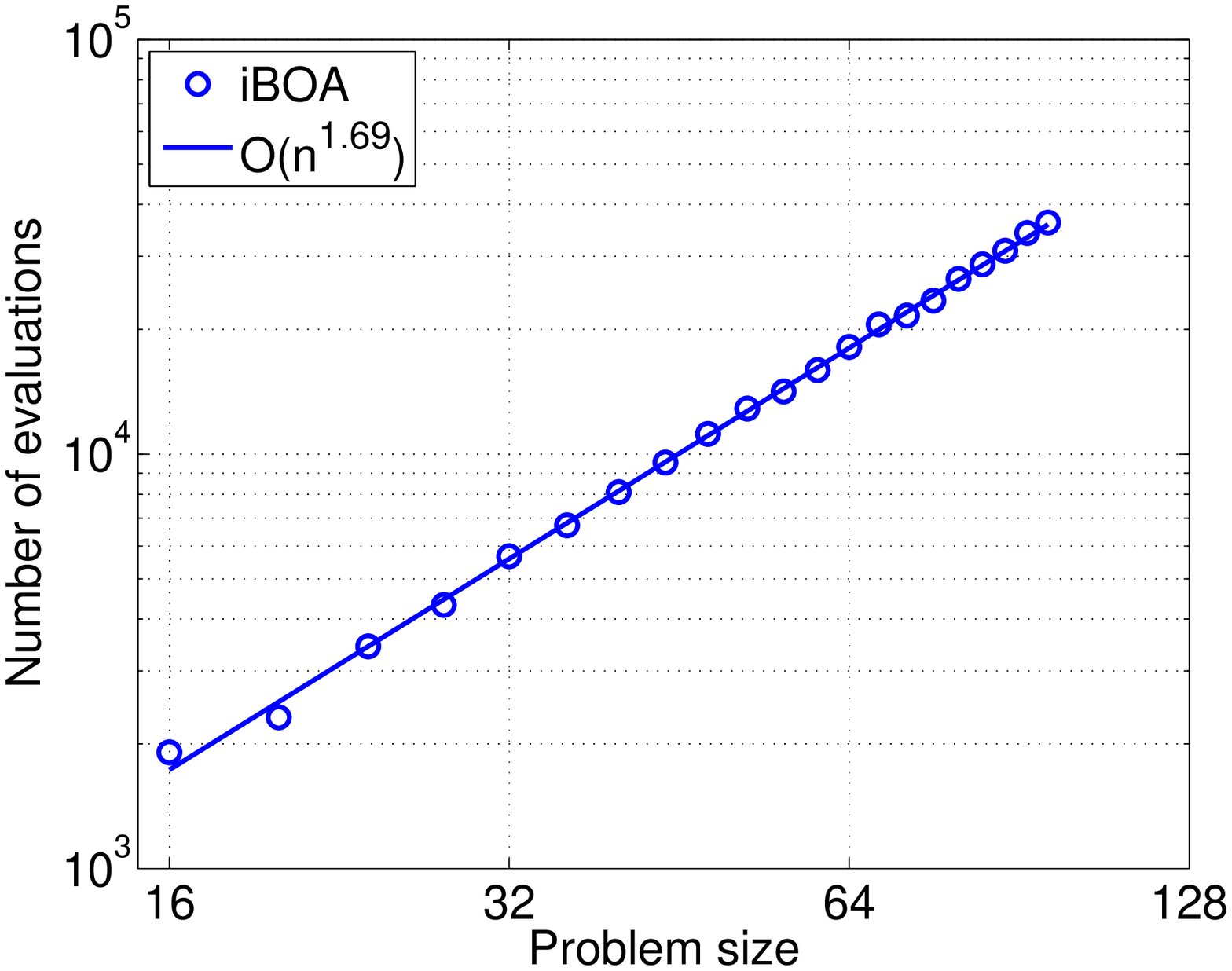}}
\subfigure[Trap-5]{\epsfig{width=0.49 \textwidth,file=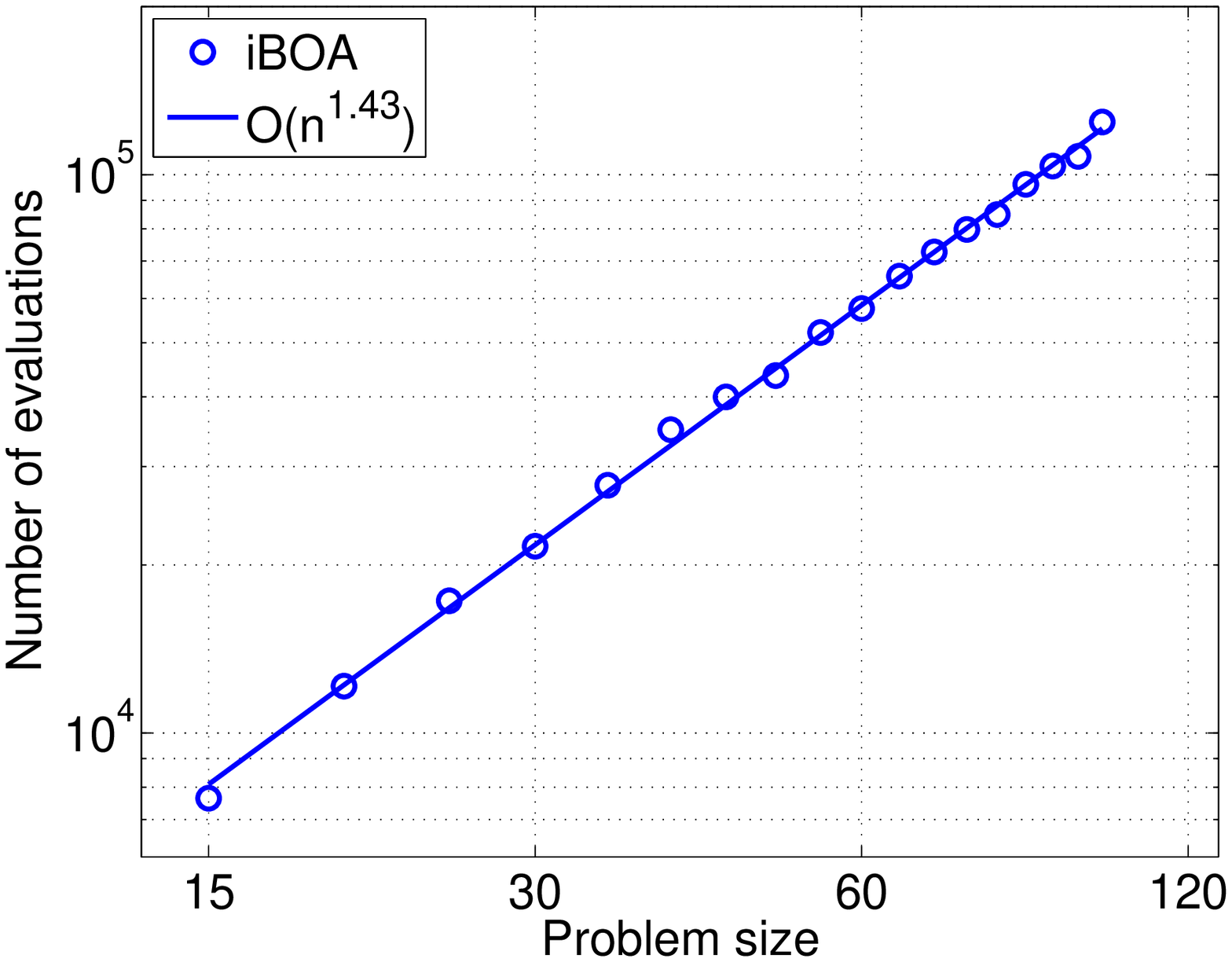}}
\caption{Performance of iBOA on concatenated traps of order 4 and 5.}
\label{fig-iboa-traps}
\end{figure}

Figure~\ref{fig-iboa-traps} shows the number of evaluations required by iBOA to reach the global optimum on concatenated traps of order 4 and 5. In both cases, the number of evaluations grows as a low-order polynomial; for trap-4, the growth can be approximated as $O(n^{1.69})$, whereas for trap-5, the growth can be approximated as $O(n^{1.43})$. While the fact that the number of evaluations required by iBOA scales worse on trap-5 than on trap-4 seems somewhat surprising, both cases are relatively close to the bound predicted by BOA scalability theory~\cite{Pelikan:02a,Pelikan:book}, which estimates the growth as $O(n^{1.55})$. 

The low-order polynomial performance of iBOA on trap-4 and trap-5 provides strong empirical evidence that iBOA is capable of finding an adequate problem decomposition because models that would fail to capture the most important dependencies on the fully deceptive problems trap-4 and trap-5 would fail to solve these problems scalably~\cite{Bosman:99,Thierens:99,Goldberg:book}. 

Figure~\ref{fig-iboa-traps} shows the number of evaluations required by standard BOA to reach the global optimum on concatenated traps of order 4 and 5. In both cases, the number of evaluations grows as a low-order polynomial; for trap-4, the growth can be approximated as $O(n^{1.69})$, whereas for trap-5, the growth can be approximated as $O(n^{2.04})$. In both cases, we see that BOA performs worse than predicted by scalability theory~\cite{Pelikan:02a,Pelikan:book}, which is most likely because of using an elitist replacement strategy, which significantly alleviates the necessity of having accurate models in the first few iterations~\cite{Pelikan:book}, and because of the potential for too strong pressure towards overly simple models due to the use of BIC metric to score network structures. In any case, we can conclude that iBOA not only keeps up with standard BOA, but without an elitist replacement strategy or niching, it even outperforms BOA with respect to the order of growth of the number of function evaluations with problem size. 

\begin{figure}[t]
\subfigure[Trap-4]{\epsfig{width=0.49 \textwidth,file=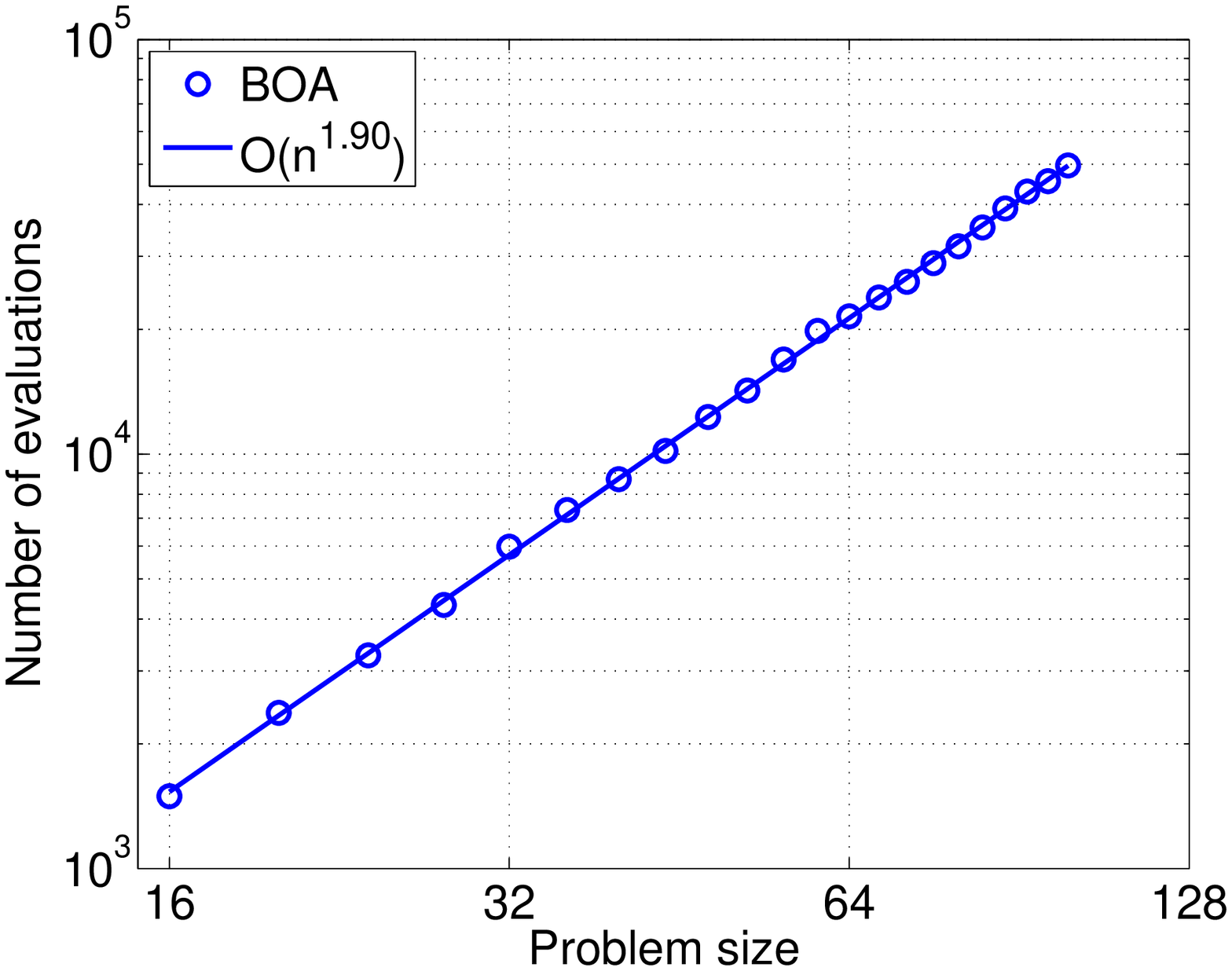}}
\subfigure[Trap-5]{\epsfig{width=0.49 \textwidth,file=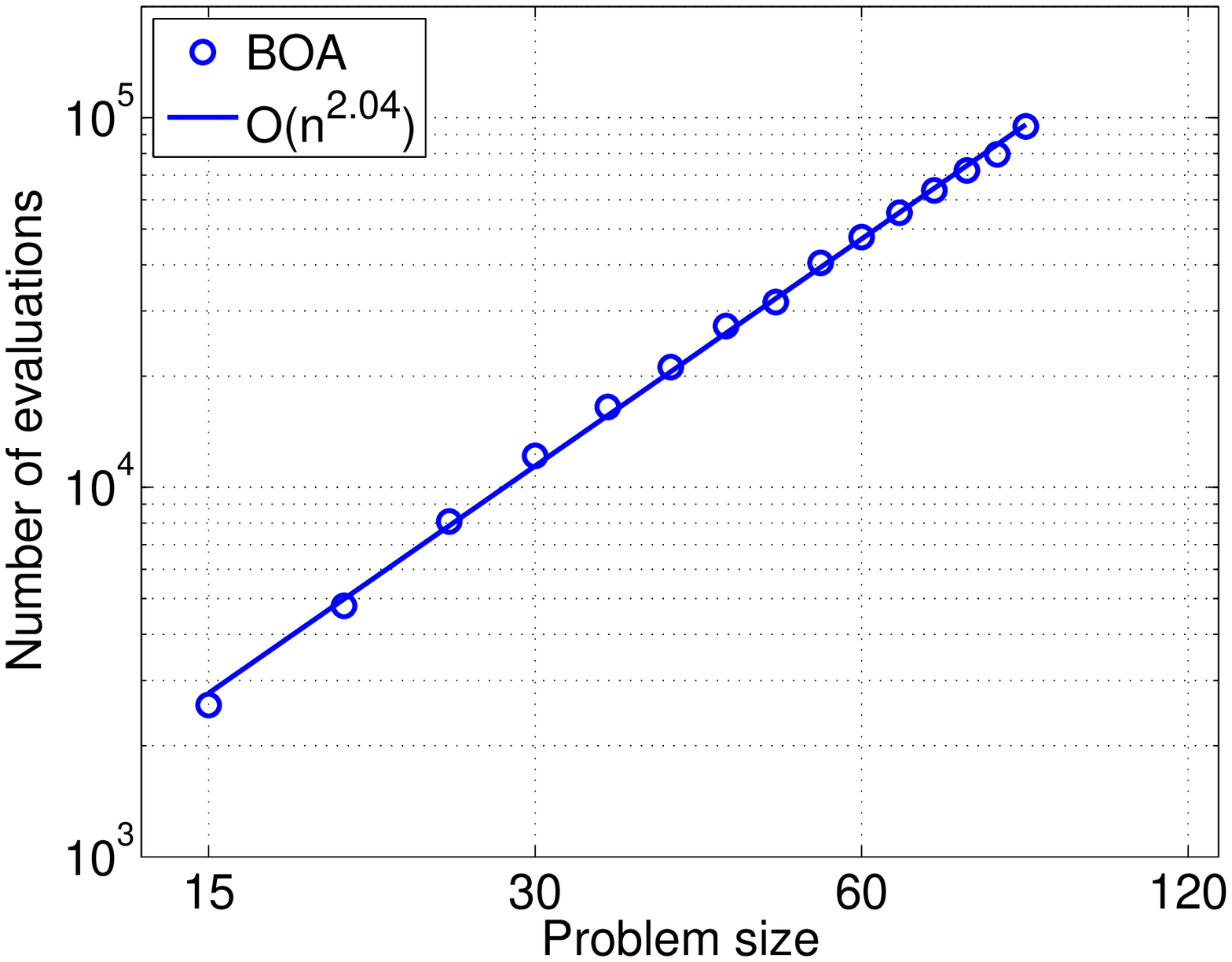}}
\caption{Performance of  standard BOA on concatenated traps of order 4 and 5.}
\label{fig-boa-traps}
\end{figure}

%==============================================================

\section{Future Work}
\label{section-future-work}
While the experiments confirmed that iBOA is capable of learning multivariate models incrementally without using a population of candidate solutions, the advantages of doing this are counterbalanced by the disadvantages. Most importantly, there are two issues that need to be addressed in future work: (1) Complexity of model representation should be improved using local structures in Bayesian networks, such as default tables~\cite{Friedman:99} or decision trees/graphs~\cite{Chickering:97,Friedman:99}. (2) Elitist and diversity-preservation techniques should be incorporated into iBOA to improve its performance. Without addressing these difficulties, the advantages of using iBOA instead of BOA are somewhat overshadowed by the disadvantages. 

%==============================================================

\section{Summary and Conclusions}
\label{section-conclusions}
This paper proposed an incremental version of the Bayesian optimization algorithm (BOA). The proposed algorithm was called the incremental BOA (iBOA). Just like BOA, iBOA uses Bayesian networks to model promising solutions and sample the new ones. However, iBOA does not maintain an explicit population of candidate solutions; instead, iBOA performs a series of small tournaments between solutions generated from the current Bayesian network, and updates the model incrementally using the results of the tournaments. Both the structure and parameters are updated incrementally. 

The main advantage of using iBOA rather than BOA is that iBOA does not need to maintain a population of candidate solutions and its memory complexity is thus reduced compared to BOA. However, without the population, implementing elitist and diversity-preservation techniques becomes a challenge. Furthermore, memory required to store the Bayesian network remains significant and should be addressed by using local structures in Bayesian networks to represent the models more efficiently. Despite the above difficulties, this work represents the first step toward the design of competent incremental EDAs, which can build and maintain multivariate probabilistic models without using an explicit population of candidate solutions, reducing memory requirements of standard multivariate estimation of distribution algorithms.

%==============================================================

\section*{Acknowledgments}
This project was sponsored by the National Science Foundation under CAREER grant ECS-0547013, by the Air Force Office of Scientific Research, Air Force Materiel Command, USAF, under grant FA9550-06-1-0096, and by the University of Missouri in St. Louis through the High Performance Computing Collaboratory sponsored by Information Technology Services, and the Research Award and Research Board programs. 

The U.S.  Government is authorized to reproduce and distribute reprints for government purposes notwithstanding any copyright notation thereon. Any opinions, findings, and conclusions or recommendations expressed in this material are those of the authors and do not necessarily reflect the views of the National Science Foundation, the Air Force Office of Scientific Research, or the U.S. Government. Some experiments were done using the hBOA software developed by Martin Pelikan and David E. Goldberg at the University of Illinois at Urbana-Champaign and most experiments were performed on the Beowulf cluster maintained by ITS at the University of Missouri in St. Louis.

\bibliographystyle{abbrv}

\end{document}